\documentclass[10pt,twocolumn,letterpaper，table, dvipsnames]{article}

\usepackage[pagenumbers]{iccv} 

\usepackage{colortbl}     
\usepackage{booktabs}     
\usepackage{amssymb}      

\newcommand{\tablestyle}[2]{\setlength{\tabcolsep}{#1}\renewcommand{\arraystretch}{#2}\centering\footnotesize}
\usepackage{subcaption}
\usepackage{multirow}
\usepackage{caption}
\usepackage{subcaption}
\definecolor{iccvblue}{rgb}{0.21,0.49,0.74}
\usepackage[pagebackref,breaklinks,colorlinks,allcolors=iccvblue]{hyperref}

\def\paperID{12518} 
\def\confName{ICCV}
\def\confYear{2025}

\title{REPARO: Compositional 3D Assets Generation with Differentiable 3D \\Layout Alignment}
\author{
Haonan Han$^{1}$\footnotemark[1] \quad
Rui Yang$^{2}$\footnotemark[1] \quad
Huan Liao$^{1}$\footnotemark[1] \quad
Jiankai Xing$^{1}$ \quad
Zunnan Xu$^{1}$ \\
Xiaoming Yu$^{3}$ \quad
Junwei Zha$^{3}$ \quad
Xiu Li$^{1}$\footnotemark[2] \quad
Wanhua Li$^{4}$\footnotemark[2] \\
{\small
$^{1}$ Tsinghua University
$^{2}$ The University of Hong Kong
$^{3}$ Tencent
$^{4}$ Harvard University
}
}

\begin{document}
\twocolumn[{%
\renewcommand\twocolumn[1][]{#1}%
\maketitle
\begin{center}
    \centering
    \captionsetup{type=figure}     
    \vspace{-1em}
    \includegraphics[width=\textwidth]{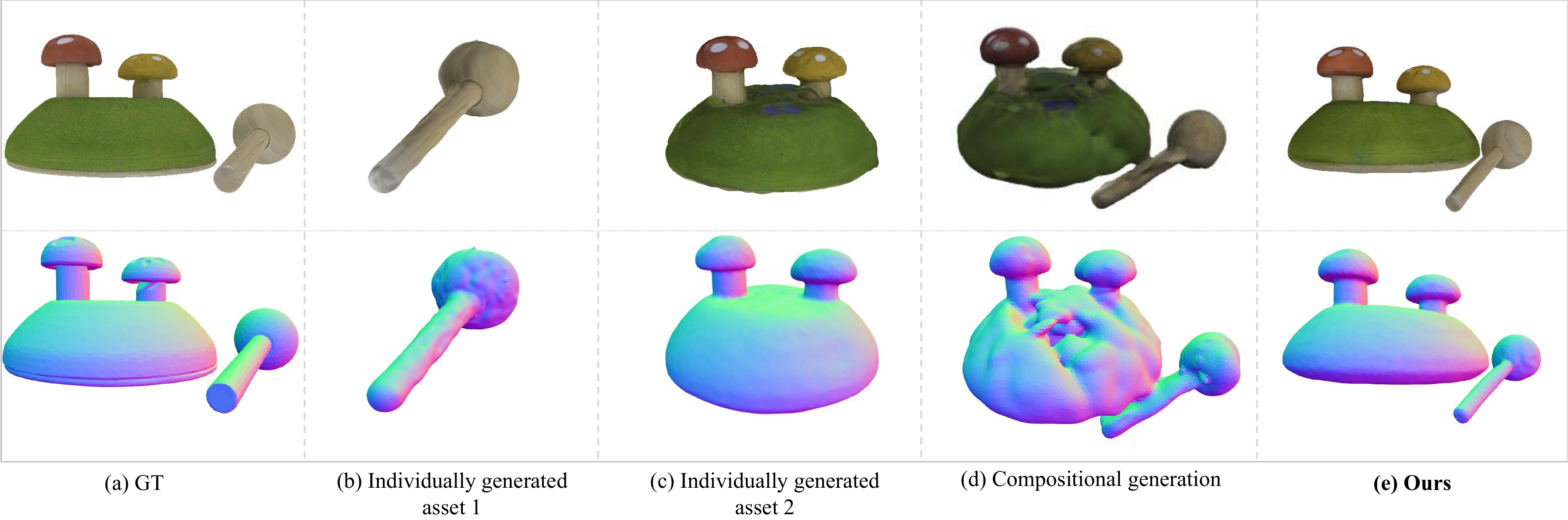}
     \vspace{-2em}
    \captionof{figure}{Qualitative comparison of generated 3D assets of the single object and multiple objects. (b) and (c) are generated 3D assets of the single object assets using DreamGaussian~\cite{tang2023dreamgaussian}; (d) is a 3D asset of multiple objects generated by DreamGaussian~\cite{tang2023dreamgaussian}; (e) is our result.
    }
    \label{fig:motivation}
\end{center}%
}]
\begin{abstract}
\let\thefootnote\relax\footnotetext{\noindent{$^*$ Equal contribution. $^\dagger$ Corresponding authors.}}
Traditional image-to-3D models often struggle with scenes containing multiple objects due to biases and occlusion complexities. To address this challenge, we present REPARO, a novel approach for compositional 3D asset generation from single images. REPARO employs a two-step process: first, it extracts individual objects from the scene and reconstructs their 3D meshes using image-to-3D models; then, it optimizes the layout of these meshes through differentiable rendering techniques, ensuring coherent scene composition. By integrating optimal transport-based long-range appearance loss term and high-level semantic loss term in the differentiable rendering, REPARO can effectively recover the layout of 3D assets. The proposed method can significantly enhance object independence, detail accuracy, and overall scene coherence. Extensive evaluation of multi-object scenes demonstrates that our REPARO offers a comprehensive approach to address the complexities of multi-object 3D scene generation from single images. The demo have been available at \href{https://reparo-3d.github.io/}{https://reparo-3d.github.io/}.
\end{abstract}
    
\section{Introduction}
\label{sec:intro}

The field of 3D content creation has advanced rapidly due to progress in reconstruction methods~\cite{nerf, kerbl3Dgaussians} and generative frameworks~\cite{poole2022dreamfusion}. Automated 3D content generation enables creators in augmented reality (AR), virtual reality (VR), gaming, and filmmaking~\cite{suzuki2022augmented, liao2024advances, li2023generative} by significantly reducing manual labor requirements.

Recent research in 3D generation primarily focuses on optimization-based 2D approaches, particularly text-to-3D~\cite{poole2022dreamfusion, Lin2023magic3d, wang2024prolificdreamer} and image-to-3D~\cite{realfusion, Tang2023makeit3d, liu2023zero123, lin2023consistent123} models. For example, Dreamfusion~\cite{poole2022dreamfusion} overcomes 3D data scarcity by distilling geometry and appearance from 2D diffusion models; and DreamGaussian~\cite{tang2023dreamgaussian} integrates 3D Gaussian Splatting into generative pipelines with mesh extraction and texture refinement.
However, despite their effectiveness in single-object scenes, these methods face challenges in reconstructing multi-object scenes. This limitation stems from dataset biases: most 3D training samples feature centrally aligned single objects, and preprocessing pipelines often re-center inputs, thereby introducing inherent positional bias. Consequently, occluded objects are frequently misrepresented as fused entities, leading to erroneously merged assets in rendered outputs, as shown in Figure~\ref{fig:motivation}~(d). 
Furthermore, typical 3D generation models output assets as monolithic mesh representations. This forces users to rely on error-prone post-processing steps to segment individual object mesh, which does not ensure the quality of the segmented assets.

To address this challenge, we propose \textbf{REPARO}, a compositional 3D generation pipeline capable of generating multiple objects from a single input image. Our approach decomposes the scene into discrete assets, then systematically reassembles and globally optimizes their spatial arrangement. This methodology not only leverages the strengths of off-the-shelf models (the ability to generate single objects with high fidelity) but also addresses inherent challenges in the multi-object scene with complex occlusions and interactions.

Specifically, our pipeline comprises two core stages. First, individual objects are isolated from the input image through cropping and inpainting to generate context-complete single-object images, which are processed using off-the-shelf image-to-3D models~\cite{dreamfusion,long2023wonder3d,TripoSR2024} to produce high-fidelity 3D assets.
Second, the 3D assets are assembled into a cohesive scene and geometrically aligned with the input image. After placing all objects within a unified coordinate system, we employ differentiable rendering to optimize the spatial arrangement of each asset. During rendering, the pose parameters (translation, rotation, and scale) of each 3D asset are treated as optimizable variables, which are iteratively refined via gradient descent using a loss function. Unlike conventional pixel-wise loss function~\cite{chen2024comboverse}, we introduce an optimal transport (OT)-based long-range appearance loss that integrates RGB color, depth, and positional information to constrain global structural relationships between objects. Therefore, this loss can match the rendered image and the reference image. Additionally, a high-level semantic loss is introduced to ensure semantic alignment and coherence between the synthesized scene and the original image context using high-level visual features.

Through these two stages, individual-object reconstruction and combined-scene optimization, our \textbf{REPARO} enhances object independence and geometric detail accuracy, particularly for occluded regions. Simultaneously, it preserves both spatial arrangement fidelity and semantic-geometric consistency across the synthesized scene. 
Critically, our method obviates the need for post hoc mesh segmentation or manual joining operations by generating instance-level meshes directly within the scene context, streamlining the 3D asset creation workflow and improving end-user usability.

To evaluate our method, we conducted experiments on 20 multi-object scenes from the Google Scanned Objects (GSO) dataset~\cite{downs2022google} (containing 2–6 objects per scene) and 20 additional in-the-wild images for qualitative analysis. Quantitative and qualitative comparisons demonstrate that \textbf{REPARO} offers significant improvements over previous techniques in ensuring the quality of assets, managing multiple assets, and processing occlusion.
Our contributions are summarized as follows: 
\begin{enumerate}
\item A two-stage compositional generation Pipeline: We propose \textbf{REPARO}, a framework to decouple multi-object 3D generation into two stages: (1) per-object high-fidelity reconstruction via inpainting-aware cropping and off-the-shelf models, and (2) global scene optimization using novel differentiable rendering. This structure ensures object-level geometric independence while preserving scene-level contextual integrity.
\item Optimal transport-driven layout loss: We design a long-range appearance loss based on optimal transport (OT) theory, which enforces multi-modal constraints (RGB, depth, positional channels) to align 3D asset layouts with the reference image. By solving for global pixel correspondences, OT mitigates local minima issues inherent in the pixel-wise loss.
\item Semantic loss: We further introduce a semantic alignment loss that leverages high-level visual features to refine object placements. This loss ensures synthesized scenes adhere to the original image’s semantic context.
\end{enumerate}

\section{Related work}
\label{related work}

\begin{figure*}[t]
    \centering
    \includegraphics[width=\textwidth]{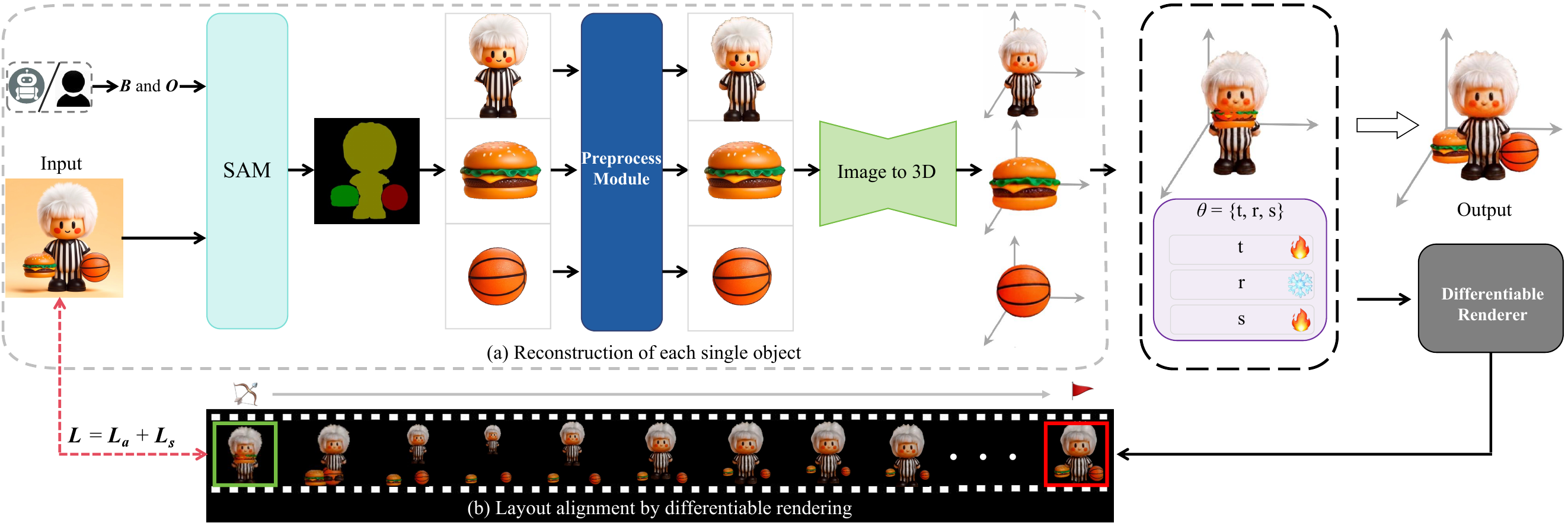}
    \caption{The diagram of the proposed REPARO. \textbf{(a)} is the pipeline to reconstruct the 3D asset of each object in the reference image. $B$ and $O$ denote bounding boxes and occlusion information of each object, respectively. If an object is occluded, the preprocessing module will complement it using the inpainting model. \textbf{(b)} is the process of layout alignment based on differentiable rendering. The parameters of reconstructed meshes are optimized by gradient descent. The loss function $L$ (Eq.~\ref{eq:total_loss}) consists of the long-range appearance loss $L_a$ and the high-level semantic loss $L_s$.}
    \label{fig:pipeline}
    \vspace{-0.5cm}
\end{figure*}

\textbf{Text-to-3D generation.}
Early exploration in 3D generation based on text~\cite{xu2023dream3d,nerf,clipmesh,clipnerf,mohammad2022clip,clip,jain2022zero} relied on a large-scale pre-trained model of text and image, CLIP~\cite{clip}, to optimize representation as the prior. In light of the promising capability of diffusion models, some works~\cite{poole2022dreamfusion,Lin2023magic3d,fantasia3d,wang2024prolificdreamer,raj2023dreambooth3d,li2024instant3d,wang2022score,seo2024let} introduce Score Distillation Sampling (SDS) to narrow the gap between novel view images rendered from 3D representation and diffusion prior. Recent works~\cite{tang2023dreamgaussian,yi2023gaussiandreamer,chen2024textto3d,li2024controllable} have used 3D Gaussian Splatting~\cite{kerbl3Dgaussians} instead of NeRF~\cite{nerf} for 3D generation, due to the fact that this efficient 3D representation shows excellent quality~\cite{yu2024mip,qin2024langsplat,li20254d} in reconstruction tasks.
 
\noindent\textbf{Image-to-3D generation.}
Differing from the ambiguity and diversity associated with textual description, 3D generation based on a single image predominantly concentrates on maintaining geometric and textural consistency between the generated assets and the input images. Some early works~\cite{realfusion,Tang2023makeit3d} used 2D diffusion models as prior knowledge to guide the training of 3D representation of the object in input images. However, with the enrichment of 3D data~\cite{deitke2022objaverse,objaverseXL}, a series of adapting fine-tuned diffusion models~\cite{liu2024one2345,shi2023zero123++,liu2023one2345++,magic123,lin2023consistent123,weng2023consistent123,zhang2023repaint123} represented by Zero-1-to-3~\cite{liu2023zero123} have been continuously developed. Subsequently, further advancements~\cite{liu2023syncdreamer,shi2023mvdream,TOSS,long2023wonder3d} have been proposed, which offer improved quality and multi-view consistency. 

\noindent\textbf{Compositional 3D generation.}
Unlike single-object reconstruction, multi-object 3D generation requires attention to both local details and global spatial relationships. Existing approaches address this through two main paradigms: (1) Single-object construction using geometric surface models \cite{Funkhouser2004modelingbyexample} and editable generative models \cite{tertikas2023generating}; (2) Scene composition employing bounding boxes \cite{po2023compositional}, text prompts \cite{wang2023luciddreaming}, Gaussian radiance fields \cite{vilesov2023cg3d}, and scene graphs \cite{gao2024graphdreamer}. Emerging approaches integrate LLMs for layout interpretation \cite{lin2023towards}, leverage 2D semantic maps \cite{yan2024frankenstein}, or combine Gaussian Splatting with LLMs \cite{yuan2024dreamscape}. Distinct from text/layout-based methods, our approach directly generates multiple high-quality 3D assets from a single compositional image while preserving input-aligned spatial arrangements.

\section{Method}
\label{method}

The field of 3D generation is advancing rapidly, with numerous image-to-3D models~\cite{tang2023dreamgaussian, TripoSR2024} capable of generating 3D assets from a single image. However, these models typically focus on individual objects, making it challenging to apply them to scenes containing multiple objects. To address this limitation, we propose \textbf{REPARO} for compositional 3D object generation from a single image. Our approach consists of two steps: (1) Extracting each target object from the given image and reconstructing their 3D meshes using off-the-shelf 3D reconstruction models. (2) Optimizing the parameters of each mesh through differentiable rendering to align their layout. In the following section, we provide a detailed explanation of these two steps.

\subsection{Reconstruction of Each Single Object}
Since most data samples in 3D datasets only have one object located centrally within the image, and most image-to-3D models recenter the object in the preprocessing step, there is an inherent center bias in these models. As a result, the generated 3D assets for single objects are better than for multiple objects. As shown in Figure~\ref{fig:motivation}, DreamGaussian~\cite{tang2023dreamgaussian} can generate the sound 3D asset for a single object, but it fails in generation for multiple objects.
Instead of reconstructing 3D assets of multiple objects at the same time, our \textbf{REPARO} first focuses on extracting individual objects from images containing several objects and then generating their 3D assets using image-to-3D models.

\noindent\textbf{Object extraction.} As illustrated in Figure~\ref{fig:pipeline}~(a), given an image $I^{ref}$ with the bounding boxes $B$ and occlusion information $O$, we use the foundation model SAM~\cite{sam} to segment each object and obtain their binary masks $M$. For occluded objects, we utilize a Stable-Diffusion-based inpainting model to reconstruct the occluded parts in the preprocessing module, thereby obtaining complete RGB priors for each object. Plus, we resize and crop the images to centralize the involved objects, which adapts to the center bias of existing image-to-3D models.

\noindent\textbf{Single object reconstruction.} After that, we employ off-the-shelf image-to-3D models, such as Dreamgaussian~\cite{tang2023dreamgaussian} and TripoSR~\cite{TripoSR2024}, to generate the corresponding 3D assets for each obtained object.

\subsection{Layout Alignment by Differentiable Rendering}
After obtaining every single object, we need a layout. To achieve the goal, first, we put every object together in one coordinate system as demonstrated in Figure~\ref{fig:pipeline}~(b), then we use differentiable rendering techniques~\cite{Laine2020diffrast,li2018differentiable,drot} to optimize the layout of each object.
Differentiable rendering aims to recover the scene parameters $\theta$ from reference image(s) $I^{ref}$ through analysis by synthesis process. Given an initial estimation of the scene, differentiable rendering can produce a rendered image $I$ together with the gradients with respect to arbitrary scene parameters $\theta$. Briefly, the gradient can be represented as:
\begin{equation}
    \frac{\partial I}{\partial \theta} = [\frac{\partial I(p_1)}{\partial \theta}, ..., \frac{\partial I(p_N)}{\partial \theta}]=[\frac{\partial I_1}{\partial \theta}, ..., \frac{\partial I_N}{\partial \theta}],
\end{equation}
where $p_i$ refers to the position of $i$-th pixel, $I(p_i)$ is the RGB color of $i$-th pixel, and $N$ is the total number of pixels. We denote $I(p_i)=I_i$ for simplicity in the following.
Combining with the loss function between rendered and reference image, we can leverage a gradient-decent-based method like Adam~\cite{Adam} to optimize the scene parameters of interest, like geometry, material, or lights. According to the chain rule, the derivative of the loss function $L(I, I^{ref})$ with respect to scene parameters $\theta$ is:
\begin{equation}
\label{eq:derivative_L_theta}
    \frac{\partial L}{\partial \theta} = \frac{\partial L}{ \partial I} \cdot \frac{\partial I}{\partial \theta}.
\end{equation}

Specifically, for our task, since we have put every generated object in the same coordinate system, the layout of objects in the initial rendered image and reference image may have quite large differences. Such a case is not suitable for the commonly used pixel-wised $L_2$ loss function and differentiable rendering methods that only compute color derivatives. Imagine that when an object in the reference image has no overlapped area with it in the rendered image, the $L_2$ loss will not change when we slightly move the object, the gradients of the loss function could get stuck in an undesired local minima and not indicate the right way for optimization.
To mitigate this issue, we propose to introduce a loss function that could find the global correspondences. For one thing, inspired by previous work~\cite{ota, drot}, we use the optimal transport algorithm to find a match between the rendered image and the reference image, thus obtaining long-range correspondences. For another thing, we utilize the feature embedding from the visual backbone to get semantic correspondences.


\subsubsection{Long-range Appearance Loss Term}

Optimal transport theory describes the following problem: suppose there are $N$ suppliers and $M$ demanders within a region. The $i$-th supplier holds $s_i$ units of goods, and the $j$-th demander needs $d_j$ units of goods. The transportation cost per unit of goods from $i$-th supplier to $j$-th demander is $c_{ij}$. The goal of the optimal transport algorithm is to find a transportation matrix $T = \{T_{ij} > 0 \mid i = 1, \ldots, N; j = 1, \ldots, M\}$ that minimizes the total transportation cost:

\begin{equation}
\label{eq:ot}
\small{
\begin{aligned}[t]
    \min_{T} & \sum_{i=1}^N \sum_{j=1}^M T_{ij} c_{ij}, \quad T_{ij} > 0, \\
    \text{s.t.} & \sum_{i=1}^{N}{T_{ij}} = d_j, 
    \quad \sum_{j=1}^{M}{T_{ij}} = s_i, 
    \quad \sum_{i=1}^{N}{s_i} = \sum_{j=1}^{M}{d_j}. 
\end{aligned}
}
\end{equation}


In our layout alignment, we consider all the pixels in the rendered image $I$ as suppliers and all the pixels in the reference image $I^{ref}$ as demanders, where $N=M$ and $s_i = d_j = 1$. 
For the cost, instead of only considering RGB color distance between $i$-th pixel and $j$-th pixel, we define it based on RGB color, depth value, and position distance at the same time:
\begin{equation}
\label{eq:cost}
    c_{ij} = \alpha \cdot \|I_i - I^{ref}_j\|_2  
            + \beta \cdot \|D_i - D^{ref}_j\|_2 
            + \gamma \cdot \|p_i-p_j\|_2,
\end{equation}
where $p_i$ and $p_j$ refer to the screen space position of $i$-th pixel and $j$-th pixel, respectively; $D$ is the depth map predicted from the rendered image $I$ using a frozen depth foundation model $F_D(\cdot)$; $D_i$ denotes the depth value of $i$-th pixel; and $\alpha$, $\beta$, and $\gamma$ are hyper-parameters. 

Based on Eq.~\ref{eq:ot} and Eq.~\ref{eq:cost}, we can obtain a transport matrix $T$ using Sinkhorn divergences~\cite{cuturi2013sinkhorn}. $T$ recording a one-to-one mapping between the rendered image $I$ and the reference image $I^{ref}$. In other words, we can find a target pixel for each pixel in the rendered image. Formally, we define $I^{ref}_{\sigma(i)}$ is the target pixel of $I_{i}$, where $\sigma(\cdot)$ is the one-to-one mapping function from the transport matrix $T$. 
In addition, rather than taking only RGB color distance as the differentiable rendering loss, we also consider RGB color, depth value, and position distance together as Eq.~\ref{eq:cost}. Accordingly, our appearance loss function can be expressed as: 
\begin{equation}
\label{eq:La}
    L_{a}(I, I^{ref}) = \frac{1}{N}\sum_{i}^N c_{i\sigma(i)}.
\end{equation}
Here, $N$ denotes the number of pixels of the rendered image, and the settings of all hyper-parameters are consistent with Eq.~\ref{eq:cost}.
To propagate the gradient of the pixel position to the scene parameter, we can define the pixel position as the projection of the shading point used for this pixel in the rasterization process as DROT~\cite{drot}.
According to Eq.~\ref{eq:derivative_L_theta} and the chain rule, the derivative of $L_a$ with respective to scene parameters $\theta$ is:
\begin{equation}
    \frac{\partial L_a}{\partial \theta} = \frac{\partial L_a}{\partial I} \cdot \frac{\partial I}{\partial \theta} +
    \frac{\partial L_a}{\partial D} \cdot \frac{\partial F_{D}}{\partial I} \cdot \frac{\partial I}{\partial \theta} + \frac{\partial L_a}{\partial p} \cdot \frac{\partial p}{\partial \theta}
\end{equation}
Note that since we freeze the depth model $F_D(\cdot)$, the gradients will totally contribute to scene parameters. As the mapping function $\sigma(\cdot)$ considers the long-range correspondence, our appearance loss term can also leverage long-range information during differentiable rendering. 

\subsubsection{High-level Semantic Loss Term}
To enhance the semantic information during the alignment process, we further propose incorporating high-level features into the loss function of differentiable rendering. Our semantic loss term is defined as:
\begin{equation}
    L_{s}(I, I^{ref}) = \frac{1}{K}\sum^{K}_{i} \| f_i - f^{ref}_i \|_2,
\end{equation}
where $f_i$ and $f^{ref}_i$ are the $i$-th embedding in the feature map $f$ and $f^{ref}$, respectively; $f$ and $f^{ref}$ are extracted using a frozen DINO-v2~\cite{dinov2} backbone $F(\cdot)$; and $K$ is the total number of embedding from the last hidden state. Based on this term, we can align the semantic relation between the rendered image and reference image from part to part. According to Eq.~\ref{eq:derivative_L_theta}, the derivative of $L_s$ with respective to layout parameters $\theta$ is:
\begin{equation}
    \frac{\partial L_s}{\partial \theta} = \frac{\partial L_s}{\partial F} \cdot
    \frac{\partial F}{\partial I} \cdot
    \frac{\partial I}{\partial \theta}
\end{equation}

So far, we have integrated the proposed loss terms together to align the layout of multiple reconstructed 3D objects:
\begin{equation}
\label{eq:total_loss}
    L(I, I^{ref}) = \lambda L_a(I, I^{ref}) + (1-\lambda) L_s(I, I^{ref}),
\end{equation}
where $\lambda$ is a hyper-parameter to adjust the weight between appearance and semantic loss term. 
In practice, we align the layout by optimizing the object's translation parameter $t=\{ \text{trans\_x}, \text{trans\_y}, \text{trans\_z} \}$ and scale parameter $s=\{ \text{scale\_x}, \text{scale\_y}, \text{scale\_z} \}$. We exclude the rotation parameter $r$ of assets from the optimization process, as assets generated via image-to-3D models are loaded into the scene with the same orientation as in the input image, eliminating the need for rotation optimization. Thanks to the proposed loss function, we can get the layout of the compositional 3D assets through differentiable rendering.



\section{Experiments}
\label{others}

\begin{figure}[t]
    \vspace{0.1cm}
    \centering
    \includegraphics[width=0.48\textwidth]{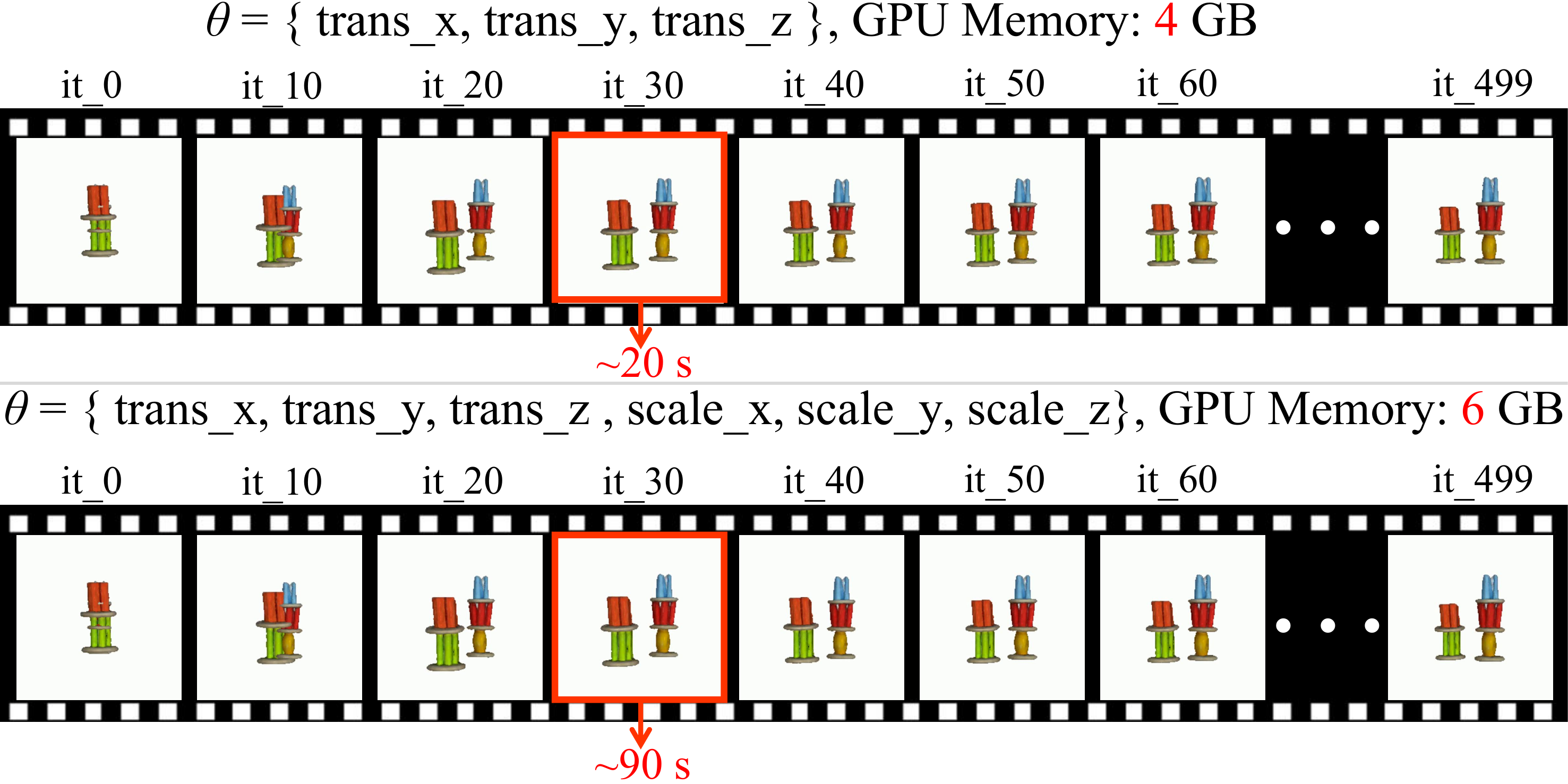}
    \vspace{-0.5cm}
    \caption{GPU memory usage and approximate elapsed time for optimization of compositional assets layout under different parameter settings.}
    \label{fig:video}
    \vspace{-0.5cm}
\end{figure}

\begin{table}[t]
    \centering
    \caption{Performance of different 3D generation models for compositional 3D assets generation. REPARO\raisebox{0.5ex}{\footnotesize$\clubsuit$} and REPARO\raisebox{0.5ex}{\footnotesize$\spadesuit$} separately denote utilizing DreamGaussian and TripoSR as the image-to-3d model based on our framework.}
    \label{table:main_res}
    \resizebox{0.48\textwidth}{!}{%
    \begin{tabular}{l|cccc}
        \toprule
        Method & CLIP $\uparrow$ & PSNR $\uparrow$ & SSIM $\uparrow$ & LPIPS $\downarrow$ \\
        \midrule
        DreamGaussian~\cite{tang2023dreamgaussian} & 0.807 & 13.280 & 0.802 & 0.240 \\
        TripoSR~\cite{TripoSR2024} & 0.795 & 17.248 & 0.863 & 0.218 \\
        Wonder3D~\cite{long2023wonder3d} & 0.801 & 13.689 & 0.807 & 0.238 \\
        LRM~\cite{hong2023lrm} & 0.812 & 13.664 & 0.806 & 0.237 \\
        \midrule
        REPARO\raisebox{0.5ex}{\footnotesize$\clubsuit$} (\textbf{ours}) & \cellcolor{gray!20}\textbf{0.833} & \cellcolor{gray!20}17.279 &\cellcolor{gray!20} 0.826 &\cellcolor{gray!20} 0.234 \\
        REPARO\raisebox{0.5ex}{\footnotesize$\spadesuit$} (\textbf{ours}) & \cellcolor{gray!20}0.822 & \cellcolor{gray!20}\textbf{17.751} & \cellcolor{gray!20}\textbf{0.865} & \cellcolor{gray!20}\textbf{0.216} \\
        \bottomrule
    \end{tabular}
    }
    \vspace{-0.2cm}
\end{table}
\begin{table}[h]
    \vspace{-0.1cm}
    \centering
    \begin{minipage}{0.23\textwidth}
        \centering
  \caption{Runtime and memory usage of each stage.}
  \vspace{-0.3cm}
  \label{tab:stage-runtime-memory}
  \tablestyle{0.3pt}{0.2}
  \resizebox{\textwidth}{!}{
  \begin{tabular}{l|cc}
    \toprule
    Stage & VRAM & Time \\
    \midrule
    SAM Segmentation       & 6 GB     & $<1s$ \\
    Inpainting             & 8 GB     & 20 $s$ \\
    Individual Generation\raisebox{0.5ex}{\footnotesize$\spadesuit$} & 6 GB     & $<1s$ \\
    Individual Generation\raisebox{0.5ex}{\footnotesize$\clubsuit$} & 8 GB     & 120 $s$ \\
    Layout Alignment       & 6 GB     & 90 $s$ \\
    \midrule
    Total\raisebox{0.5ex}{\footnotesize$\spadesuit$}       & $\leq$8GB    & 120 $s$ \\
    Total\raisebox{0.5ex}{\footnotesize$\clubsuit$}& $\leq$8GB    & 240 $s$ \\
    \bottomrule
  \end{tabular}}
    \end{minipage}
    \hfill
    \begin{minipage}{0.24\textwidth}
        \centering
    \caption{Comparison with representative methods on perceptual quality and resource usage.}
    \vspace{-0.3cm}
    \label{table:compare_gpu_time}
    \tablestyle{0.2pt}{0.2}
    \resizebox{\textwidth}{!}{
    \begin{tabular}{l|ccc}
        \toprule
        Method & CLIP $\uparrow$ & GPU & Time \\
        \midrule
        DreamGaussian & 0.807 &  8 GB  & 120 $s$ \\
        TripoSR & 0.795 & 6 GB  & $<1s$  \\
        Wonder3D & 0.801 &  16 GB & 180 $s$ \\
        LRM & 0.812 & 14 GB & 20 $s$ \\
        \midrule
        REPARO\raisebox{0.5ex}{\footnotesize$\clubsuit$} (\textbf{ours}) & \cellcolor{gray!20}0.833 & \cellcolor{gray!20}$\leq$8GB & \cellcolor{gray!20}240 $s$ \\
        REPARO\raisebox{0.5ex}{\footnotesize$\spadesuit$} (\textbf{ours}) & \cellcolor{gray!20}0.822 & \cellcolor{gray!20}$\leq$8GB & \cellcolor{gray!20}120 $s$ \\
        \bottomrule
    \end{tabular}}
    \end{minipage}
    \vspace{-0.5cm}
\end{table}
\subsection{Implementation details}
In the reconstruction of 3D assets from a single object, we utilize SAM-ViT-H~\cite{sam} as the segmentation model to obtain masks corresponding to each bounding box. In the preprocessing module, we extract images of the objects using these masks and subsequently resize and crop the images to center the objects. We then apply the Stable Diffusion-based inpainting model~\cite{stablediffusioninfinity} to address occlusions in the objects. 
During the generation of 3D assets from a single object, we employ the DreamGaussian~\cite{tang2023dreamgaussian} and TripoSR~\cite{TripoSR2024}.
For layout alignment, we use Nvdiffrast~\cite{Laine2020diffrast} as our differentiable rendering framework. During the differentiable rendering process, we optimize the translation $x, y, z$, and rotation $r$ parameters of each 3D asset using gradient descent. Specifically, we employ the Adam optimizer~\cite{Adam} with a learning rate of $0.02$ and a weight decay of $0.999$. The total number of iterations is set to $500$. 
All experiments were conducted on RTX 3090 GPUs, and as shown in Figure~\ref{fig:video}, optimizing parameter $t$ alone takes approximately 20 seconds and utilizes 4GB of memory, while optimizing both parameters $t$ and $s$ to their optimal states requires about 90 seconds and utilizes 6GB of memory.
We report per-stage runtime and memory in Table~\ref{tab:stage-runtime-memory}. All modules run on $\leq$8GB VRAM, covering SAM, inpainting, 3D generation, and layout alignment. Table~\ref{table:compare_gpu_time} compares our method with baselines in runtime and memory. REPARO achieves better performance under similar or lower resource budgets.

To validate the effectiveness of our method, we selected 20 samples containing multiple objects from the Google Scanned Objects (GSO) dataset~\cite{GSO} as our test set. We used images from 18 views to compute quantitative metrics, including CLIP score~\cite{clip}, Peak signal-to-noise ratio (PSNR), the structural similarity (SSIM)~\cite{ssim}, and Perceptual Similarity (LPIPS)~\cite{lpips}. Moreover, we also conducted a subjective evaluation of the generation quality for different models. We collected opinions from 40 participants. Participants selected the visually best subjective quality option among the generation results of 4 models for the same input image. We computed the preference score for each model by dividing the selection count of a specific model by the total selections, reflecting the human preference distribution for each model.

\begin{figure*}[t]
    \vspace{0.6cm}
    \centering
    \moveleft 0.68cm \hbox{ 
        \includegraphics[width=1\textwidth]{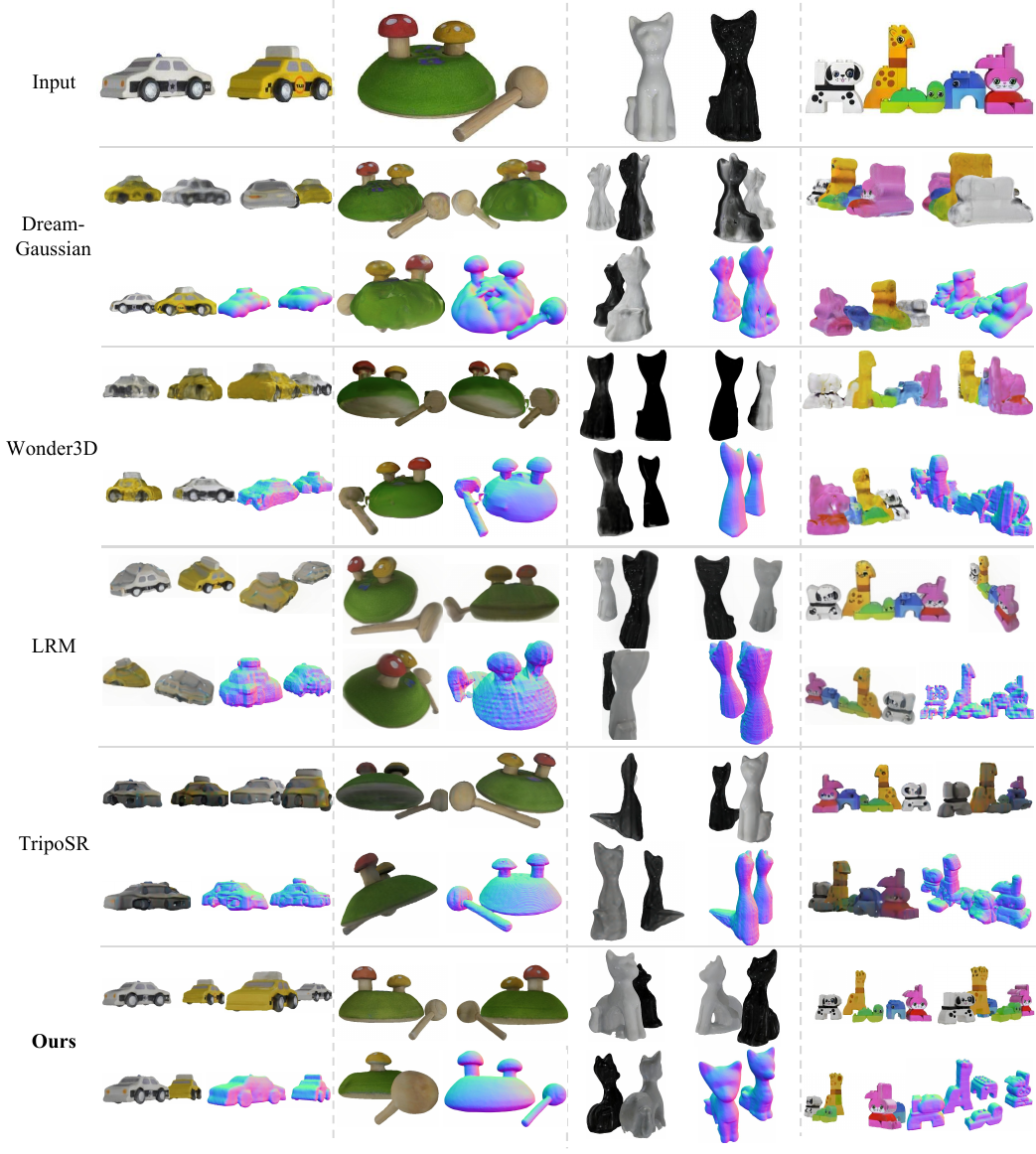}
    }
    \caption{Qualitative comparison with different image-to-3D generation models. Given an input image, previous methods produce inaccurate textures and geometry with noticeable artifacts. Our method generates high-quality, high-fidelity compositional assets with the correct spatial layout.}
    \label{fig:SOTA_ablation}
\end{figure*}

\begin{table*}[!ht]
    \centering
    \begin{minipage}{0.3\textwidth}
        \centering
        \tablestyle{1.5pt}{0.9}
        \raisebox{-\height}{\parbox{\textwidth}{
            \caption{Comparison of different loss functions.}
            \label{tab:different_loss_term}
            \begin{subtable}{\textwidth}
                \centering
                \caption{DreamGaussian-based REPARO.}
                \begin{tabular}{cc|cccc}
                    \toprule
                    $L_a$ & $L_s$ & CLIP $\uparrow$ & PSNR $\uparrow$ & SSIM $\uparrow$ & LPIPS $\downarrow$ \\
                    \midrule
                    \checkmark &    & 0.833 & 17.296 & 0.826 & 0.233 \\
                              & \checkmark  & 0.809 & 17.820 & 0.849 & 0.210 \\
                    \checkmark & \checkmark  & \cellcolor{gray!20}0.833 & \cellcolor{gray!20}17.279 & \cellcolor{gray!20}0.826 & \cellcolor{gray!20}0.234 \\
                    \bottomrule
                \end{tabular}
            \end{subtable}
            
            \vspace{0.2cm}
            \begin{subtable}{\textwidth}
                \centering
                \caption{TripoSR-based REPARO.}
                \begin{tabular}{cc|cccc}
                    \toprule
                    $L_a$ & $L_s$ & CLIP $\uparrow$ & PSNR $\uparrow$ & SSIM $\uparrow$ & LPIPS $\downarrow$ \\
                    \midrule
                    \checkmark &    & 0.822 & 17.765 & 0.865 & 0.216 \\
                              & \checkmark  & 0.813 & 17.906 & 0.867 & 0.212 \\
                    \checkmark & \checkmark  & \cellcolor{gray!20}0.822 & \cellcolor{gray!20}17.751 & \cellcolor{gray!20}0.865 & \cellcolor{gray!20}0.216 \\
                    \bottomrule
                \end{tabular}
            \end{subtable}
        }}
    \end{minipage}
    \hfill
    \begin{minipage}{0.32\textwidth}
        \centering
        \tablestyle{1.5pt}{0.9}
        \raisebox{-\height}{\parbox{\textwidth}{
            \caption{Comparison of optimization parameters.}
            \label{tab:ablation_theta}
            \begin{subtable}{\textwidth}
                \centering
                \caption{DreamGaussian-based REPARO.}
                \begin{tabular}{ccc|cccc}
                    \toprule
                    T &  R &  S & CLIP $\uparrow$ & PSNR $\uparrow$ & SSIM $\uparrow$ & LPIPS $\downarrow$ \\
                    \midrule
                      \checkmark & &     & 0.832 & \textbf{17.369} & 0.826 & \textbf{0.232} \\
                       \checkmark&\checkmark &   & 0.826 & 16.906 & 0.820 & 0.243 \\
                       \checkmark& &  \checkmark & \cellcolor{gray!20}\textbf{0.833} & \cellcolor{gray!20}17.279 & \cellcolor{gray!20}\textbf{0.826} & \cellcolor{gray!20}0.234 \\
                       \checkmark&\checkmark & \checkmark  & 0.831 & 17.122 & 0.823 & 0.241 \\
                    \bottomrule
                \end{tabular}
            \end{subtable}
            
            \begin{subtable}{\textwidth}
                \centering
                \caption{TripoSR-based REPARO.}
                \begin{tabular}{ccc|cccc}
                    \toprule
                    T &  R &  S & CLIP $\uparrow$ & PSNR $\uparrow$ & SSIM $\uparrow$ & LPIPS $\downarrow$ \\
                    \midrule
                      \checkmark & &     & 0.819 & \textbf{17.797} & 0.865 & 0.216 \\
                       \checkmark&\checkmark &   & 0.822 & 17.768 & 0.864 & 0.219 \\
                       \checkmark& &  \checkmark & \cellcolor{gray!20}0.822 & \cellcolor{gray!20}17.751 & \cellcolor{gray!20}\textbf{0.865} & \cellcolor{gray!20}\textbf{0.216} \\
                       \checkmark&\checkmark & \checkmark  & \textbf{0.826} & 17.752 & 0.864 & 0.219 \\
                    \bottomrule
                \end{tabular}
            \end{subtable}
        }}
    \end{minipage}
    \hfill
    \begin{minipage}{0.35\textwidth}
        \centering
        \tablestyle{1.5pt}{0.9}
        \raisebox{-\height}{\parbox{\textwidth}{
            \caption{Comparison of information in appearance loss $L_a$.}
            \label{tab:ablation_La}
            \begin{subtable}{\textwidth}
                \centering
                \caption{DreamGaussian-based REPARO.}
                \begin{tabular}{lc|cccc}
                    \toprule
                    Info in $L_a$ &  OT & CLIP $\uparrow$ & PSNR $\uparrow$ & SSIM $\uparrow$ & LPIPS $\downarrow$ \\
                    \midrule
                    RGB    & $\times$    & 0.800 & 18.054 & 0.853 & 0.206 \\
                    RGB    & \checkmark  & 0.828 & 17.009 & 0.821 & 0.245 \\
                    RGBXY  & \checkmark  & 0.828 & 17.069 & 0.821 & 0.243 \\
                    RGBDXY & \checkmark  & \cellcolor{gray!20}0.833 & \cellcolor{gray!20}17.279 & \cellcolor{gray!20}0.826 & \cellcolor{gray!20}0.234 \\
                    \bottomrule
                \end{tabular}
            \end{subtable}
            
            \begin{subtable}{\textwidth}
                \centering
                \caption{TripoSR-based REPARO.}
                \begin{tabular}{lc|cccc}
                    \toprule
                    Info in $L_a$ & OT & CLIP $\uparrow$ & PSNR $\uparrow$ & SSIM $\uparrow$ & LPIPS $\downarrow$ \\
                    \midrule
                    RGB    & $\times$    & 0.774 & 17.918 & 0.868 & 0.216 \\
                    RGB    & \checkmark  & 0.819 & 17.837 & 0.865 & 0.215 \\
                    RGBXY  & \checkmark  & 0.823 & 17.847 & 0.865 & 0.213 \\
                    RGBDXY & \checkmark  & \cellcolor{gray!20}0.822 & \cellcolor{gray!20}17.751 & \cellcolor{gray!20}0.865 & \cellcolor{gray!20}0.216 \\
                    \bottomrule
                \end{tabular}
            \end{subtable}
        }}
    \end{minipage}
    \vspace{0cm}
\end{table*}

\subsection{Main results}

\textbf{Quantitative experiment.} Table~\ref{table:main_res} compares the performance (CLIP score, PSNR, SSIM, and LPIPS) of various methods for reconstructing 3D assets of multiple objects using one reference image. DreamGaussian-based REPARO and Triposr-based REPARO, our proposed methods, show significant improvements over baseline methods (DreamGaussian~\cite{tang2023dreamgaussian}, Wonder3D~\cite{long2023wonder3d}, and LRM~\cite{hong2023lrm}). DreamGaussian-based REPARO achieves 83.3\% CLIP score, 17.279 PSNR, 0.826 SSIM, and 0.234 LPIPS, indicating better alignment between reconstructed 3D assets and ground-truth 3D assets. Triposr-based REPARO exhibits superior performance in PSNR (17.751) and SSIM (0.865) and the lowest LPIPS (0.216), highlighting its exceptional quality in image reconstruction and perceptual similarity. These results show that better reconstruction quality on single objects is beneficial for structural similarity and perceptual similarity during differentiable rendering.

\noindent\textbf{Qualitative experiment.} Some qualitative comparison results are shown in Figure~\ref{fig:SOTA_ablation}, where our Triposr-based REPARO can produce high-quality 3D assets with multiple objects. By contrast, DreamGaussian~\cite{tang2023dreamgaussian}, Wonder3D~\cite{long2023wonder3d}, and LRM~\cite{hong2023lrm} have problems in layout and completeness, since they possess preferences on an individual object.


\noindent\textbf{User study.} In addition to numerical metrics, we conducted a user study to compare our method with others, gathering responses from participants who assessed the realism of different 3D assets. Our approach received 61\% approval, while Dreamgaussian received 19\% approval, LRM received 12\% approval, and Wonder 3D received 8\% approval. The results consistently showed that our approach was favored over previous methods. For more detailed results, please refer to the supplementary material.




\subsection{Ablation study}

\begin{figure}[t]
    \centering
    \includegraphics[width =0.48\textwidth]{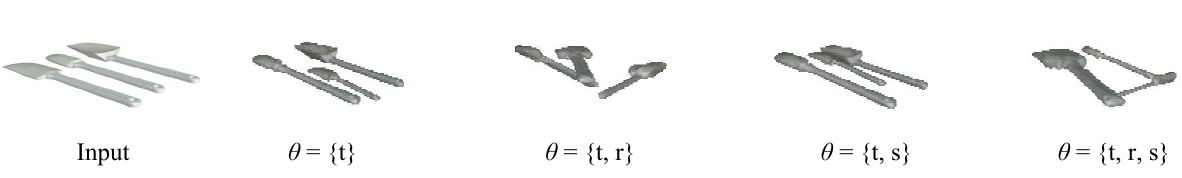}
    \caption{Qualitative comparison for different options of optimization parameters $\theta$.}
    \vspace{0cm}
    \label{fig:op_param}
\end{figure}


\begin{figure}[t]
    \centering
    \includegraphics[width =0.45\textwidth]{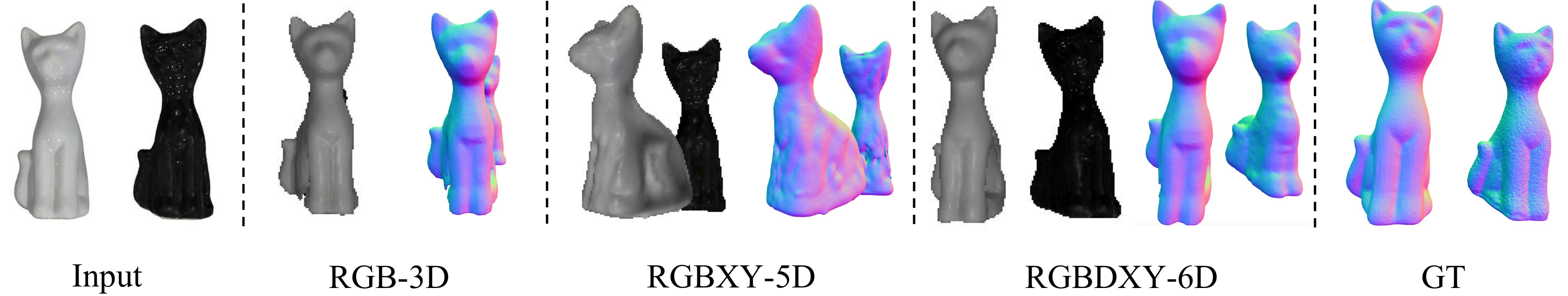}
    \caption{Qualitative comparison for different terms in the appearance loss $L_a$ (Eq.~\ref{eq:La}).}
    \vspace{-0.3cm}
    \label{fig:ablation_inverse_render}
\end{figure}

\begin{figure*}[t]
    \centering
    \includegraphics[width=\textwidth]{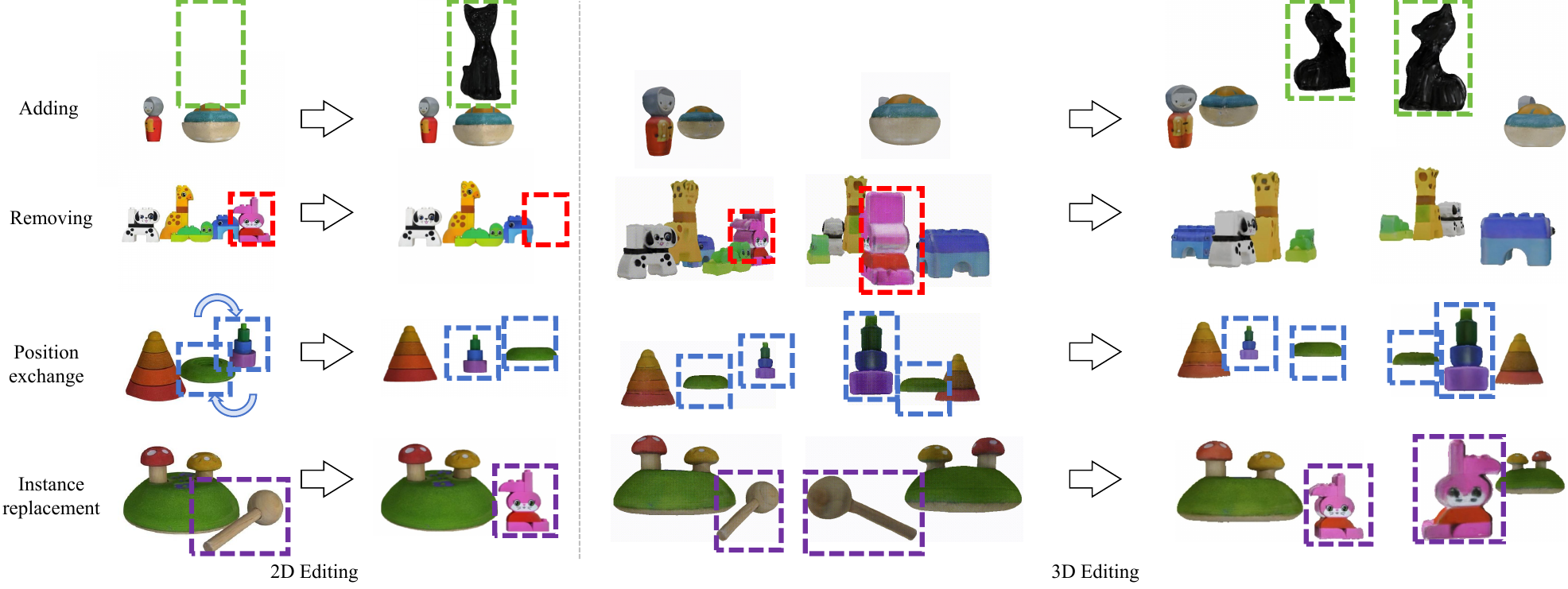}
    \caption{3D scene editing based on the implementation of 2D image editing. Based on our REPARO framework, it is possible to implement four types of 3D editing guided by 2D image editing, namely adding, removing, position exchanging and instance replacement.}
    \label{fig:3D_Edit}
    \vspace{-0.5cm}
\end{figure*}

\noindent\textbf{Ablation for different loss functions in differentiable rendering.} 
We conducted ablation experiments on different loss components in Eq.~\ref{eq:total_loss}, with the results presented in Table~\ref{tab:different_loss_term}. The results indicate that incorporating the high-level semantic loss function $L_s$ can bring improvements in both SSIM and LPIPS. However, the enhancement in CLIP score is not as pronounced. Additionally, when only $L_s$ is used, there is a notable decrease in CLIP score. For instance, in the DreamGaussian-based model, the CLIP score decreased by $2.1\%$. This suggests that relying solely on feature supervision can reduce the effectiveness of layout alignment.

\noindent\textbf{Ablation for different options of scene parameters $\theta$.} Table~\ref{tab:ablation_theta} provides an ablation study on the impact of various parameter optimization settings within $\theta$ across both DreamGaussian-based and TripoSR-based REPARO models. Specifically, we compare combinations of translation ($t$), rotation ($r$), and scale ($s$) optimizations. For DreamGaussian-based REPARO, the combination of parameters $t$ and $s$ achieves the highest CLIP score (0.833), competitive PSNR (17.279), and the lowest LPIPS (0.234), indicating a high level of perceptual quality and fidelity. Similarly, for TripoSR-based REPARO, optimizing $t$ and $s$ yields a balance between perceptual similarity (CLIP 0.822) and structural fidelity, with the lowest LPIPS (0.216) and strong SSIM (0.865), aligning closely with the input image. 

\noindent Notably, as shown in Figure~\ref{fig:op_param}, including the rotation parameter $r$ in the optimization tends to degrade alignment with the input image orientation, as shown by the rotation-inclusive configurations, which generally exhibit slight reductions in CLIP scores and increased LPIPS. As a result, we select the combination of $t$ and $s$ for the final configuration, as it achieves optimal perceptual alignment without sacrificing the fidelity to the original image orientation.

\noindent\textbf{Ablation for long-range appearance loss term.} As shown in Eq.~\ref{eq:La}, long-range appearance loss $L_a$ incorporates RGB color distance, position distance, and depth value distance. More importantly, we use the optimal transport to compute pixel matching from a global perspective, thereby obtaining long-range correspondences. Quantitative results in Table~\ref{tab:ablation_La} demonstrate that in the DreamGaussian-based model, introducing long-range correspondence results in a 2.8\% improvement in CLIP score; in the TripoSR-based model, it results in a 4.5\% improvement in CLIP score. Additionally, when position distance and depth distance are introduced separately, there is no significant change in quantitative results, but there is a noticeable improvement in qualitative results, as illustrated in \Cref{fig:ablation_inverse_render}.

\begin{table}[t]
    \centering
    \tablestyle{3.0pt}{0.9}
    \caption{DreamGaussian-based ($\clubsuit$), TripoSR-based ($\spadesuit$), and GT-based ($\blacklozenge$) results.}
    \label{table:main_res}
    \begin{tabular}{l|cccc}
        \toprule
        Method & $\text{CLIP}_{\text{s}}$ $\uparrow$ & PSNR $\uparrow$ & SSIM $\uparrow$ & LPIPS $\downarrow$ \\
        \midrule
        REPARO$^\clubsuit$ & 0.833 & 17.279 & 0.826 & 0.234 \\
        REPARO$^\spadesuit$ & 0.822 & 17.751 & 0.865 & 0.216 \\
        REPARO$^\blacklozenge$ & 0.906 & 15.708 & 0.789 & 0.321 \\
        \bottomrule
    \end{tabular}
    \vspace{-0.3cm}
\end{table}

\noindent \textbf{The impact of 3D asset quality.}
As shown in \cref{table:main_res}, the experimental results indicate that the process of generating individual assets in the first stage introduces significant errors.
Therefore, we still need a model capable of generating high-quality 3D assets.

\subsection{Extension: Using 2D editing to guide 3D editing}

We further explored potential applications of our combinatorial generation framework. As depicted in the \Cref{fig:3D_Edit}, the entire framework relies on a single input image as guidance, enabling both the reconstruction of individual objects and layout alignment through differentiable rendering. Leveraging this capability, we integrated pre-existing image editing techniques to facilitate editing of the generated 3D scenes. Specifically, we demonstrate four types of editing operations: adding, removing, position exchange, and instance replacement. The 2D editing process is implemented using the AnyDoor~\cite{chen2024anydoorzeroshotobjectlevelimage} model. By comparing the 3D scenes before and after editing, it is evident that the REPARO framework exhibits robust spatial arrangement capabilities and maintains consistency for each assets.

\section{Conclusion}

In conclusion, we introduce REPARO, a comprehensive approach for generating compositional 3D scenes from single images. Our REPARO addresses the multi-object issue by decomposing the scene into individual objects, reconstructing their 3D meshes, and optimizing their layout through differentiable rendering. During differentiable rendering, we incorporate an optimal transport-based long-range appearance loss which considers RGBDXY information, and a high-level semantic loss which aligns the feature correspondence. Consequently, REPARO is able to obtain multi-object 3D assets with visual-spatial arrangement and contextual consistency. One limitation of our method is that severe occlusion may lead to hallucinated parts inconsistent with the image, and in semantically implausible scenes, layout optimization tends to favor common-sense interpretations over the actual image. 
We hope our method can pave the way for image to 3D generation in multi-object scenes. 

{
    \clearpage
    \newpage
    \small
    \newpage
    \bibliographystyle{ieeenat_fullname}
    \bibliography{main}
}

\end{document}